# 3DGS-LSR：Large-Scale Relocation for Autonomous Driving Based on 3D Gaussian Splatting

Haitao Lu[1,2], Haijier Chen[1,2], Haoze Liu[1,2], Shoujian Zhang[1,*], Bo Xu[1], Ziao Liu[1]

[1] The School of Geodesy and Geomatics, Wuhan University, No. 129 Luoyu Road, Wuhan 430079, People's Republic of China

E-mail: shjzhang@sgg.whu.edu.cn

**Abstract**

In autonomous robotic systems, precise localization is a prerequisite for safe navigation. However, in complex urban environments, GNSS positioning often suffers from signal occlusion and multipath effects, leading to unreliable absolute positioning. Traditional mapping approaches are constrained by storage requirements and computational inefficiency, limiting their applicability to resource-constrained robotic platforms. To address these challenges, we propose 3DGS-LSR: a large-scale relocalization framework leveraging 3D Gaussian Splatting (3DGS), enabling centimeter-level positioning using only a single monocular RGB image on the client side. We combine multi-sensor data to construct high-accuracy 3DGS maps in large outdoor scenes, while the robot-side localization requires just a standard camera input. Using SuperPoint and SuperGlue for feature extraction and matching, our core innovation is an iterative optimization strategy that refines localization results through step-by-step rendering, making it suitable for real-time autonomous navigation. Experimental validation on the KITTI dataset demonstrates our 3DGS-LSR achieves average positioning accuracies of 0.026m, 0.029m, and 0.081m in town roads, boulevard roads, and traffic-dense highways respectively, significantly outperforming other representative methods while requiring only monocular RGB input. This approach provides autonomous robots with reliable localization capabilities even in challenging urban environments where GNSS fails.

## 1. Introduction

Accurate and reliable localization in complex environments is essential for enabling unmanned technologies, such as autonomous driving. Current localization methods primarily depend on sensors like GNSS, LiDAR, IMUs, and cameras. However,GNSS-based localization encounters significant challenges in urban areas [1]. Tall buildings can obstruct satellite signals, leading to unstable reception, while the multipath effect — caused by signals reflecting off building surfaces — introduces positioning errors. When GNSS fails to provide accurate positioning, the trajectories and positions obtained from IMUs, LiDAR, and cameras can accumulate errors over time [2,3,4].To overcome these challenges, high-precision maps that have been pre-collected are often used for localization in complex scenarios where GNSS fails to provide accurate positioning. These maps contain essential information about absolute positions and orientations, which aids in localization [5]. Specifically, environmental feature data, along with their corresponding high-precision absolute positions and orientations, is gathered in advance. When the carrier returns to the environment, it can retrieve and match its sensor data with the pre-stored environmental feature data to accurately determine its current absolute position and orientation [6].

---

[2] These authors contributed equally to this work as co-corresponding authors.

[*] Authors to whom any correspondence should be addressed.

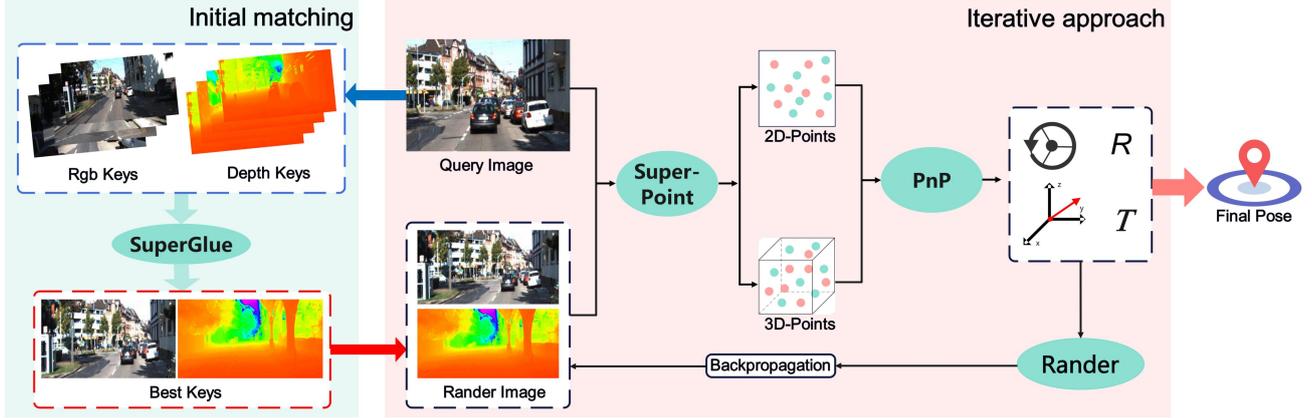

**Figure 1** A Visual Relocation System for High-Precision Pose Estimation：The system first matches the images captured by the camera with a reference image database to identify the reference image with the highest similarity, which is determined as the coarse pose.Subsequently, feature points between the target image and the reference image are extracted and matched using the RGB and depth images under the coarse pose, establishing a 2D-3D relationship. The pose of the target image is computed using the Perspective-n-Point (PnP) algorithm, resulting in a relatively accurate pose. Finally, RGB and depth images are rendered under the relatively accurate pose, and an iterative refinement method is employed to determine the final refined pose of the target image.

These maps are available in various formats, including point cloud maps, vector maps, and bird's-eye view (BEV) maps. Each format has its limitations when used for absolute positional assistance in practical navigation scenarios [7]. Point cloud maps enable high-precision environmental alignment by matching LiDAR data; however, they are computationally demanding and require significant storage capacity. Vector maps utilize road networks and landmark information for large-scale navigation, but their positioning accuracy may not always be sufficient, and the landmark information can become outdated over time [8]. BEV maps provide a comprehensive representation of the environment, making it easier to match with camera images. However, limitations in resolution and field of view can negatively affect matching performance and positioning accuracy [9].

3D Gaussian(3DGS) maps represent a state-of-the-art method for mapping environments, providing high-fidelity visual representations from various perspectives while meticulously preserving texture details. Furthermore, their rapid rendering capabilities make them ideal for meeting the real-time demands of autonomous driving applications.

We introduce a new relocation method based on 3D Gaussian maps.First,creating a 3D Gaussian map specifically designed for autonomous driving scenarios. Next, we use RGB images captured by the onboard camera to match them with RGB images rendered from the Gaussian map. Once the initial matching is complete, we combine the RGB image with the depth image to estimate the vehicle's approximate position. This approximate position is then used to let the 3DGS map to render the image from the corresponding viewpoint. By iteratively repeating this process, our method achieves high-speed, centimeter-level repositioning with excellent accuracy.

In this paper, we propose a high-precision relocation method based on 3D Gaussian maps, highlighting the following key contributions:

（1）By combining efficient matching algorithms with the rapid rendering capabilities of Gaussian maps, this method allows for real-time, iterative repositioning with high accuracy.

（2）The method removes reliance on unreliable GNSS for absolute positioning in urban environments. Instead, vehicles can achieve centimeter-level repositioning using only camera sensors.

Our proposed method was evaluated on various sequences from the KITTI dataset, which represent different scenarios, and successfully demonstrated its real-time performance and localization accuracy.

## 2.RELATED WORK

In the field of relocation, existing methods mainly include fingerprint relocation based on radio signals and relocation based on sensors, such as LIDAR sensors, cameras, and so on.

Radio signal-based relocation methods utilize signals such as Wi-Fi and Bluetooth. While these methods perform well in indoor environments, they are susceptible to interference, which poses some challenges [10]. These challenges include the need for frequent updates to the signal fingerprint information, limited accuracy in relocation, and a restricted application range primarily suited for small indoor areas. Consequently, these methods are not suitable for large-scale outdoor scenarios, such as automated driving [11].

Sensor-based relocation can be categorized into two main types: LiDAR-based relocation technology and camera-based relocation technology. Camera-based visual relocation has several significant advantages over LiDAR and other high-cost equipment.

Firstly, visual relocation can be achieved using ordinary cameras, resulting in lower costs for sensors, easier deployment, and reduced maintenance expenses compared to LiDAR. Secondly, cameras not only capture geometric information but also provide rich texture and illumination data, which enhances scene understanding and localization in complex environments.

Various research methodologies exist for visual relocation, including feature matching-based methods, scene coordinate regression [12], pose regression [13], and direct image alignment [14]. Notably, 2D-3D feature-matching methods are predominant in this domain. These methodologies ascertain the camera's pose—comprising both position and orientation—by extracting feature points from two-dimensional images and aligning them with pre-constructed three-dimensional scene feature points. The construction of 3D scene models usually relies on Structure-from-Motion (SfM) [15], Truncated Signed-Distance Function (TS) based on the range image, and the Signed-Distance Function (TSDF) [16] based on range images or LiDAR image building techniques [17]. However, these traditional methods have deficiencies in the ability to express lighting and texture information, and the generated models are mainly used for geometric reconstruction, which is difficult to support image rendering at new viewpoints, limiting their applicability in complex scenes.

As technology advances, the methods used to represent maps for relocation are also changing. Unlike traditional fixed formats such as vector maps or point cloud maps, newer representations like Neural Radiance Fields (NeRF) [18] and 3D Gaussian Splatting offer a more detailed depiction of scene information [19,20,21]. These innovative methods can generate high-fidelity images from any viewpoint, providing new opportunities for research in relocation based on these advanced map representations.

The introduction of Neural Radiation Fields (NeRF) and subsequent research has established a novel paradigm for viewpoint synthesis and camera poseestimation. NeRF is capable of providing precise and realistic representations of static scenes characterized by complex geometries and lighting conditions. It achieves this by modeling the scene as a continuous three-dimensional volume and learning the inherent properties of its radiation field. However, despite the high accuracy of NeRF-based camera pose estimation methods, they encounter several limitations. For example, the computational complexity associated with the inverse projection and light projection processes of the NeRF model results in prolonged training and inference times, which complicates the achievement of real-time performance. Furthermore, NeRF's limited capacity to accommodate dynamic scenes further restricts its applicability in specific scenarios.

In contrast, the 3D Gaussian map represents an advanced method of map representation that offers enhanced computational efficiency and superior dynamic scene representation [22,23]. This technique facilitates rapid rendering and allows for the seamless integration of multimodal data, making it particularly advantageous for real-time applications, such as autonomous driving. The SplatLoc [24] approach was the first to implement the 3DGS technique for visual localization by introducing 3D Gaussian primitives to represent scenes. This method has demonstrated performance that either exceeds or is comparable to leading implicit visual localization techniques in terms of rendering and localization efficacy. Subsequently, GauLoc [25] put forth an implicit feature metric alignment method that exhibits increased accuracy and robustness in complex scenarios. The GSLoc [26] method realized efficient image alignment and visual localization, particularly in texture-free environments, by employing a coarse-to-fine optimization strategy and integrating a fuzzy kernel to alleviate the non-convex optimization challenges. Likewise, GsplatLoc [27] achieved efficient image alignment and visual localization in texture-free contexts by minimizing the disparity between the rendered depth map and the observed depth map, thereby attaining exceptionally accurate camera position estimation.

Current methods for position estimation using 3D Gaussian maps primarily focus on small indoor environments. However, these methods encounter challenges when applied to large-scale outdoor open scenes. To address this issue, the approach proposed in this paper utilizes 3DGS maps and camera sensors to achieve low-cost, high-precision, and rapid repositioning for users in extensive outdoor environments.

## 3.OUR APPROACH

In this section, we elaborate on our system, which is divided into three categories: 3DGS map rendering, enhanced feature extraction and matching, and refined visual relocation. The overall system framework is shown in Fig. 1.

*3.1 3D Gaussian map construction*

For the large-scale outdoor scenes faced by autonomous driving, due to the influence of a single viewing angle, high contrast, and an increase in the number of moving objects, the traditional 3DGS image construction method will produce a blind spot in the viewing angle, which affects the acquisition of depth information and the three-dimensional sense of the object, and also leads to overexposure or the loss of shadow details in the image, resulting in the phenomenon of artifacts, which reduces the accuracy of the three-dimensional reconstruction. Therefore, we adopt OmniRe [28], a "panoramic" system capable of handling diverse participants, to render the image, which integrates laser data and image information, effectively adapts to the limitation of a single viewpoint, provides more credible depth information, and builds a clear and blur-free image. At the same time, by constructing a dynamic neural scene graph, using Gaussian representation to model various dynamic actors in the scene, and assigning dedicated local normative space to different dynamic entities, we achieve comprehensive capture and reconstruction of complex dynamic scenes, restore the scene with high fidelity, ensure the completeness of the details, and satisfy the demand for high-precision 3DGS map models required by our high-precision localization for autonomous driving.

Among them, in order to realize the effective representation of various dynamic objects, the Gaussian scene graph is composed of sky nodes, background nodes, rigid nodes, and non-rigid nodes; the background nodes are represented by a set of static Gaussians $G_{bg}$, which are initialized by accumulating LIDAR points and randomly generated extra points [29]; the rigid nodes are represented by $\bar{G}_v$ in the local space, which are in turn transformed into the world space by the following formula:

$$G_v = T_v(t) \otimes \bar{G}_v \qquad (1)$$

The non-rigid nodes are further subdivided into two categories: the SMPL nodes for walking or running pedestrians as well as the deformable nodes for out-of-distribution non-rigid instances, and the two types of nodes are represented in the world space formulated as follows:

$$G_{SMPL}(t) = T_h(t) \otimes LBS(\theta(t), \bar{G}_{SMPL}) \qquad (2)$$

$$G_{deform}(t) = T_h(t) \otimes (\bar{G}_{deform} \otimes F_\ell(\bar{G}_{deform}, e_h, t)) \qquad (3)$$

The three nodes are transformed into a Gaussian representation in world space. This Gaussian is then rendered and stitched together by a rasterizer. Meanwhile, the sky node is depicted using an optimized environment texture map and is rendered separately. The rendering results $C_{sky}$ from the sky node are then combined with the overall rendering results of the other nodes to produce the final output, defined by the following formula [29,30]:

$$C = C_G + (1 - O_G)C_{sky} \qquad (4)$$

Optimize all of the above parameters simultaneously in one stage to achieve the best overall rendering, the optimization function is defined as follows:

$$\zeta = (1-\lambda_r)\zeta_1 + \lambda_r\zeta_{SSIM} + \lambda_{depth}\zeta_{depth} + \lambda_{opacity}\zeta_{opacity} + \zeta_{reg} \qquad (5)$$

The $\zeta_1$ is the L1 loss used to measure the pixel-level difference between the rendered image and the target image, $\zeta_{SSIM}$ is the structural similarity loss of the rendered image, $\zeta_{depth}$ is the difference between the rendered Gaussian depth values and the sparse depth signal from the LiDAR data, and denotes the regularization constraints applied to the different Gaussian representations.

### 3.2 Enhanced feature point matching method

The front-end task of our whole system is feature point extraction and matching, common feature point extraction and matching methods, such as SIFT+FLANN [31], ORB+BFMatcher [32], etc. The effect is limited by feature selection, there is the possibility of information loss, and more sensitive to noise and occlusion, although the speed of processing speed is faster, but in terms of matching accuracy is not accurate enough, so it is difficult to cope with the complex visual environment of large-scale scenes faced by automated driving, and it is impossible to find the best matching image in the sparse gallery and accurately find a sufficient number of feature matching points. Therefore, our system adopts the SuperPoint+SuperGlue [33,34] combination of feature extraction and matching, using SuperPoint to extract feature points and descriptors in the image, and SuperGlue to realize the matching between images, organically combining the three steps of feature point detection, feature description, and feature matching based on deep learning into a complete deep network architecture, so as to realize an end-to-end deep network architecture. architecture, thus realizing an end-to-end image feature point matching processing front-end module [35]. The method has good performance in dealing with low-quality images caused by fast vehicle movement and missing or obstructed light in the field of view, and at the same time, the number of extracted feature points and the accuracy of the matching results are much better than the traditional methods when facing large-scale changing images.

SuperPoint is a deep network trained by a self-supervised approach to recognize a large number of feature points and also generate high-dimensional fixed-length descriptors [36]. The SuperPoint network mainly consists of three parts: encoder, feature point decoder, and descriptor decoder. In this network, the feature point detector and descriptor in the encoding phase share the same encoding network; while in the decoding phase, the According to the requirements of specific tasks, feature point decoding and descriptor decoding adopt different structures, so as to learn their independent network parameters, the specific algorithms are shown in the following Algorithm1 [33].

**Algorithm1: SuperPoint**

**Input:** An image of size $H \times W$

**Output:** Feature point probabilities, descriptors

**1.** Extraction of base features using the VGG-style encoder while reducing the image scale

**2.** The feature point decoder converts the dimensions of the image output by the encoder into a tensor of the size of $H/8 \times W/8 \times 64$, Then processed by Softmax and Reshape module, a new $H \times W$ size tensor can be obtained, and after the encoder can finally get the probability that each pixel point is a feature point.

**3.** The descriptor decoder first acquires a semi-dense descriptor and then transforms the size of the image output by the encoder into a tensor of $H/8 \times W/8 \times D$ size, The rest of the description is obtained by double cubic polynomial interpolation, and then the normalized descriptor processing module is introduced to finally obtain the tensor of $H \times W \times D$ size, and the local descriptor of each pixel is finally obtained by the descriptor decoder.

SuperGlue is a novel graph neural network algorithm [37] that can efficiently filter out outliers during feature matching. It can utilize the differentiable optimal transmission theory for feature matching, and combines the attention mechanism with the introduction of two-dimensional feature points and aggregation strategy, so that the algorithm can efficiently process feature points and descriptors obtained from both traditional and deep learning methods to generate accurate matching pairs. The specific algorithm flow is shown in the following Algorithm2 [34]:

**Algorithm2: SuperGlue**

**Input:** Detected feature points and descriptors

**Output:** An accurate match

**1.** Using the GNN (graph neural network) module, the feature points and their descriptors are converted into feature matching vectors, and the self-attention and cross-attention mechanisms in the attention mechanism continuously reinforce the (L=7) vector F;

**2.** At the optimal matching level, the inner product is used to obtain the score matrix, and after the Sinkhorn algorithm (T iterations) processes these points to solve the optimal feature assignment matrix and obtain the overall accurate matching results of the final image feature points.

*3.3 High-precision attitude convergence*

1）Initial Matching to Determine the Coarse Position: when determining the initial position of the target location, the feature matching between the target query image and the RGB images in the reference image library is first performed by the SuperGlue algorithm in the reference image library, and the feature matching is optimized by using the spatial context and the global features, to identify the reference RGB image $I_{GS}$ with the highest degree of similarity, The reference library here is the multiple RGB images generated from the 3DGS map. When we obtain the 6D position parameters of the reference image, it is used as the initial coarse position of the target point.

2）PnP to determine the accurate positional attitude: in the process of determining the coarse positional attitude is matched to the reference image, which contains the RGB image and depth image generated by the 3DGS map, which for each spatial point we can obtain a set of three-dimensional coordinates under the camera coordinate system, i.e., ($X_G$、$Y_G$、$Z_G$) through the chi-square coordinates of the spatial point with the feature points obtained by the projection of the normalized planar chi-square coordinates, we adopt the Perspective-n-Point (PnP) method to obtain the relative accurate position of the target image location based on the initial rendering reference map obtained.

EPnP (efficient perspective n point) algorithm is an efficient way to solve the PnP conditional problem [38]; the method is small in computation and the complexity is , for the extraction of a larger number of feature points, has a faster computational efficiency. The core theory of the EPnP algorithm is to use the non-coplanar virtual control point linearly weighted representation of an arbitrary sign point in the camera coordinate system, so it is generally the first to determine the virtual control point and sign point in the camera coordinate system position, construct the relationship between the world and the camera coordinate system is defined as follows:

$$\begin{cases} P_i^k = \sum_{j=1}^{4} a_{ij} C_j^k \ (i=1,2,3\cdots,n) \\ P_i^m = \sum_{j=1}^{4} a_{ij} C_j^m \ (i=1,2,3\cdots,n) \end{cases} \quad (6)$$

$P_i^k$ and $P_i^m$ are the positions of the marker points in the world coordinate system and the corresponding camera coordinate system, $C_j^k$ and $C_j^m$ are the positions of the virtual control points in the world coordinate system and the corresponding camera coordinate system, respectively, and are the weighting coefficients corresponding to each marker point, which sums up to 1. In order to obtain the positions of the virtual control points in the camera coordinate system, the construction equations are defined as follows:

$$Z_c \begin{bmatrix} u_i \\ v_i \\ 1 \end{bmatrix} = K \begin{bmatrix} R & T \\ 0 & 1 \end{bmatrix} \begin{bmatrix} X_i^m \\ Y_i^m \\ Z_i^m \\ 1 \end{bmatrix} \qquad (7)$$

Where $K$ is the internal reference matrix of the camera and $C_j^m = \begin{bmatrix} X_i^m & Y_i^m & Z_i^m & 1 \end{bmatrix}^T$ are the unknown quantities to be solved, i.e. the coordinates of the virtual control point in the camera coordinate system. The 6D position of the target point is then solved using the ICP (Iterative Closest Point) absolute localization method [39].

In order to make use of more effective information and reduce the influence of noise between matching feature points, we proceed to construct a least squares optimization problem to adjust the estimates (Bundle Adjustment) [40], by constructing a least squares problem for the pixel and spatial point position relation matrix error to minimize it Li algebraically, the formula is defined as follows:

$$T^* = \arg\min_T \frac{1}{2} \sum_{i=1}^{n} \| u_i - \frac{1}{s_i} K T P_i \|_2^2 \cdot \qquad (8)$$

Find the optimal solution as the exact solution for the 6D position of the target point。

3）Iterative calculations to determine the refined position：

Considering the constraints of positioning accuracy under one computation, in order to further improve the accuracy, we iteratively refine the accurate positioning determined by one PnP, and refer to the following algorithm 3 for details.

The accurate positioning information is re-inputted to the 3DGS map, and the rendering generates the RGB image and depth image under the accurate position, and the latest rendering result and the target map are utilized to iteratively execute the positioning Determination, at the end of each execution a new target point pose is generated, the iterative process aims at determining a more accurate target point pose, and the latest result obtained after satisfying the iteration conditions is used as the final refined pose, i.e., it is considered to be the pose information of the target point.The detailed relocation iterative optimization structure is as follows: Algorithm 3.

## 4. Experimental Evaluation

In this section, we begin by introducing the dataset utilized for our study and outlining the implementation details of our method, along with the performance evaluation metrics. We then assess our proposed method in terms of its feature-matching capabilities and localization performance.

**Algorithm3**: Iterative optimization

**Input**: relatively accurate pose $P_1$

**Output**: fine pose $P_{truth}$

// determine an upper limit for the number of cycles $i_{max}$

$i_{max} \leq 10$

for $i \leftarrow 1$ to $i_{max}$ do
    //3DGS rendering
    $(C_{rgb}^i, C_{depth}^i, C_{pose}^i) = C_{rander}(P_i)$
    // PnP algorithm processes the target and renders the data
    $P_{i+1} = C_{PnP}((C_{rgb}^i, C_{depth}^i, C_{pose}^i), C_{objective})$
    $\cdots\cdots i = (1, i_{max})$
    // loop termination condition
    **If** $|P_n - P_{n-1}| \leq 0.01$ **then**
        $P_{truth} = P_n$
    **Else**
        execute $C_{rander}()$ and $C_{PnP}()$ function
end

$P_{truth} = P_n^{end}$

The evaluation demonstrates the superiority of our approach through comparative tests with various high-performance methods. The purpose of these test experiments is to validate the effectiveness of our proposed method, which aims to achieve fast and efficient re-localization based on 3DGS maps and visual information in large-scale autonomous driving scenarios

### 4.1 Experimental Setup

**Datasets** We used the Karlsruhe Institute of Technology and Toyota Technological Institute dataset(KITTI) dataset for 3DGS mapping as well as relocation tests to evaluate our approach.The KITTI dataset is a standard dataset developed by the Karlsruhe Institute of Technology and the Toyota Technological Institute in Chicago for a wide range of applications in autonomous driving research. The dataset contains a rich set of real-world driving scenarios, in which we use the camera, IMU inertial navigation device, Velodyne HDL-64E S2 LIDAR sensor data, and the MASK information of the objects to complete the construction of high-precision 3DGS maps of large-scale scenarios, and then we use a single image provided by the camera to complete the relocation, and the relocation of the true value of the position of the carrier is based on the dataset. The positional truth value of the repositioning is calculated based on the carrier positional truth value provided by the dataset and the external reference matrix of the camera.

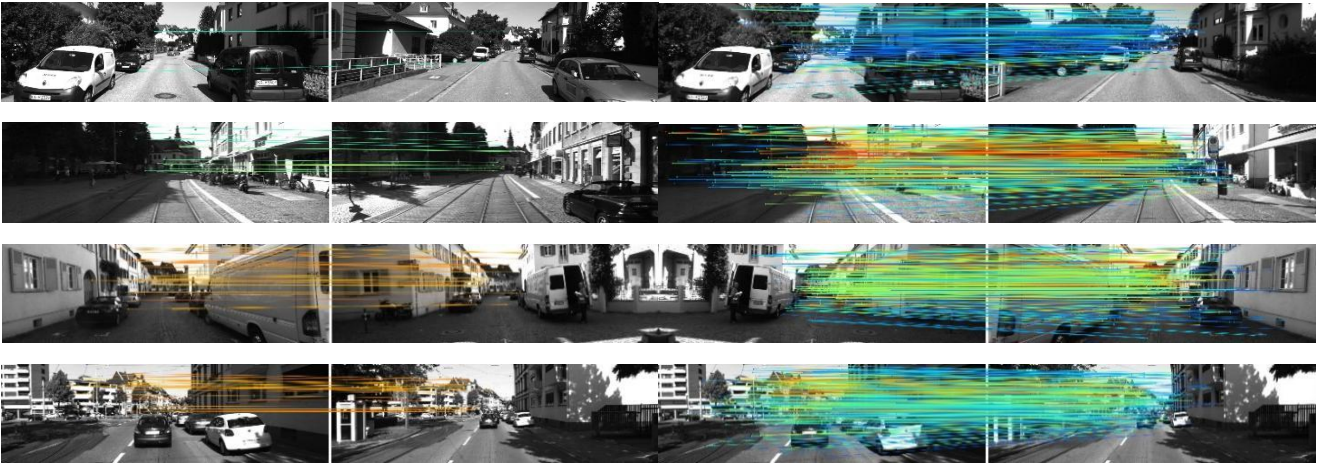

**Figure 2** Comparison of feature matching effect. A comparison of the matching results using the traditional method and superglue in different scenarios is given in the figure. Superglue achieves better-matching results in each scenario.

**Implementation Details** Our localization process was implemented on a system with an Intel Core i9-13620H CPU, 16 GB of RAM, and an NVIDIA RTX 4090 GPU with 24 GB of graphics memory. The algorithm was developed in Python and PyTorch, utilizing a custom CUDA kernel to accelerate the rasterization and backpropagation processes inherent in our micro-renderable approach. This configuration ensures that our approach achieves real-time performance, which is critical for real-world applications of retargeting systems.

*4.2 Image feature point matching comparison*

In our approach, a portion of 2D RGB maps and their corresponding depth maps are first rendered based on the established outdoor large-scale 3DGS maps as the base map library for the initial matching of the relocation query maps. To ensure the lightness of the base map library and the timeliness of the query image matching, the base map library has to be rendered with a certain degree of sparseness, i.e., only one image will be rendered after a certain distance of being rendered densely. Instead of rendering it densely.

However, in large-scale outdoor scenes, the sparse rendering of images results in significant differences in the viewing angle between the query image and even the best-matching RGB image in the gallery. Therefore, we must ensure that the query can accurately select the best matching image in the base gallery, even in the presence of large viewing angle differences. We adopts SuperPoint+SuperGlue combination method in the front-end feature point extraction and matching part, which performs well in the outdoor scene, especially in the scene with blurred image and insufficient light, and also can dynamically adjust its operation rate according to the complexity of the matching task, which greatly improves the number of feature points extracted and the accuracy of the matching result comparing with the traditional method.

Figure 2 gives a comparison of the feature extraction and matching effects of the traditional method (ORB+Brief) and our method (SuperPoint+SuperGlue) under multiple scenarios. Obviously, the feature point extraction and matching ability of the combined method is significantly better than the traditional method in large-scale scenarios. The first and second rows indicate the road environments with sparse features and the road environments with large steering angles, respectively, in which the combined ORB + Brief method suffers from matching errors in these two challenging environments. The third and fourth rows represent road environments with more features, and the SuperPoint + SuperGlue combination method can extract more high-quality matches to ensure the accuracy of the initial displacements calculated by pnp.

In addition to this, with the 3DGS iteration by iteration, the feature matching effect between the rendered image and the query image is also gradually improved, as shown in Fig. 3. In the iteration-by-iteration process; it can be found that the view angle gap between the new image rendered by the 3DGS map and the query image is gradually reduced, and the number of feature points matched between the images is gradually improved, which makes the accuracy of calculating the relative displacement also gradually improved, which is in line with the iterative progression of the relocation process.

Figure 3 visually reflects the improvement in feature point extraction and matching as the number of iterations increases. Table 1 quantitatively demonstrates the improvement effect of iterations on feature point extraction and matching using indicators such as the number of feature point matches, the average confidence of feature matches, and the uniformity of feature point distribution. The number of feature point matches reflects the number of feature points in two images. The average confidence of feature matches indicates the reliability of all feature matches. The uniformity of feature point distribution is the difference between the normalized standard deviation of feature point distribution and 1; the closer it is to 1, the more uniformly the feature points are distributed across the two images.

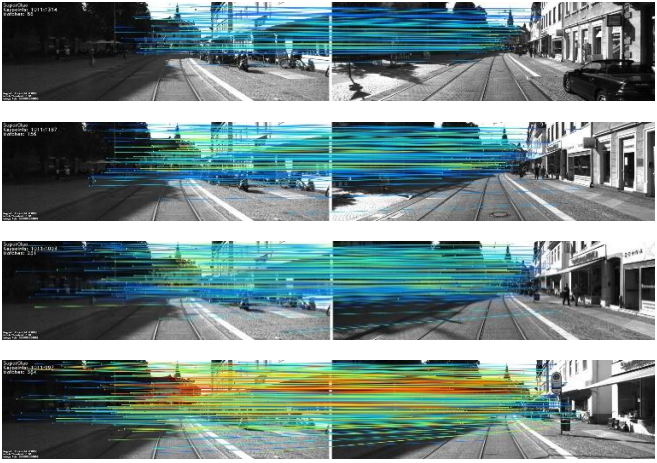

**Figure 3:** Iterative matching graph. As the iteration proceeds, the rendered image gets closer and closer to the query graph, and the number of feature point matches and the confidence of feature point matches between the two gradually increase

**Table 1:** trajectory errors in different scenarios

| Number of Iterations | Number of Feature Point Matches | Average Confidence of Feature Matches | Uniformity of Feature Point Distribution |
|---|---|---|---|
| 1 | 58 | 0.735 | 0.354 |
| 2 | 126 | 0.826 | 0.466 |
| 3 | 257 | 0.892 | 0.598 |
| 4 | 383 | 0.957 | 0.701 |

*4.3 Relocation experiments*

We tested the relocation accuracy of our method in three typical scenarios in the KITTI dataset, namely, the town road, the boulevard road, and the vehicle-dense highway.

The distribution of the error between the relocation results and the reference truth value of some pictures in a selected sequence of each scene is counted as shown in the following figure 4.

Seq1, Seq2, and Seq3 in the above figure represent the localization error statistics on the town road, boulevard road, and traffic-intensive highways, respectively. Among them, the localization performance was best in the town road, while it was slightly weaker on boulevard roads. This may be due to the comparison shown in Figure 5, between the first and second rows. When rendering the complex and variable light and shadow scenes in the forest, compared to town roads, there is a certain degree of blurring in the randered images, as indicated by the red box in the right image of the second row.

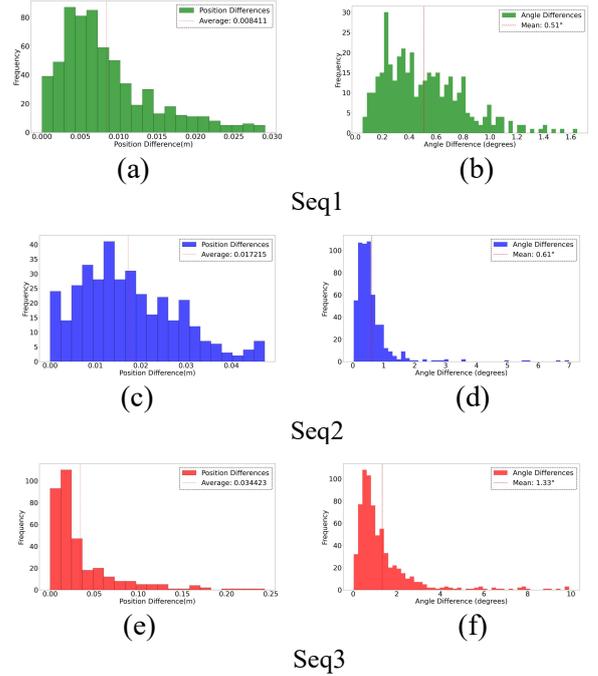

**Figure 4:** Histogram of position and angle localization errors in town roads, boulevard roads, and traffic-dense highway scenes

This affects the extraction of feature points and the accuracy of subsequent relative localization calculations. In the vehicle-intensive highway scene, the average positional error and average angular error of the relocation calculation are relatively poorer among the three scenes, which may be because, although our 3DGS construction method effectively avoids the ghosting effect of rendering dynamic objects, in feature-degraded environments like highways, there are fewer feature point matches between images, as shown in the third row of Figure 5, with poor distribution uniformity. This results in a decrease in the final average error.

Then we randomly selected a period of time in each of the three sequences respectively, each with a length of 300 meters. and arranged the pictures taken by the camera sensors in the vehicle trajectory during that time in order, and relocated them sequentially as query pictures, and statistically analyzed the obtained trajectory results.

The absolute trajectory errors of the localization results and the reference truth in each of the three scenarios are synthesized statistics, including the root mean square error, standard deviation, mean, median, minimum, and maximum values as shown in Table 2.

From the Table 2, it can be seen that the RMSE and Std value of the absolute trajectory error of the town road are the smallest, respectively 0.026935 and 0.018506, but the Mean value of the trajectory error of the boulevard road is the smallest, 0.01312, and the statistics of each of the vehicle-dense road absolute trajectory error are all on the large side, so that the RMSE in the town road scenario as well as in the boulevard scene is close to 0.02, and the accuracy is the The best, although the effect is slightly worse under the highway scene, but the RMSE is also close to 0.07, combined with the above trajectory diagram can be seen, the three scenes, the overall have higher positioning accuracy results. At the same time, the standard deviation of the errors in the three scenarios are 0.018506, 0.023182, 0.069639 respectively, which shows that the absolute trajectory error is very robust, and in summary, it shows that our system still meets the demand for high-precision localization in the case of continuous operation of vehicles.

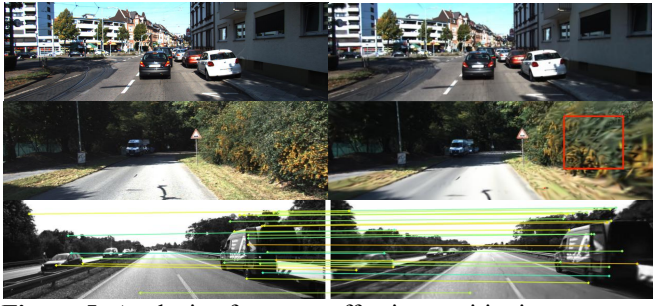

**Figure 5:** Analysis of reasons affecting positioning accuracy

**Table 2:** trajectory errors in different scenarios(m)

| Seq | RMSE | Std | Mean | Median | Min | Max |
|---|---|---|---|---|---|---|
| 1 | 0.026 | 0.018 | 0.019 | 0.015 | 0.003 | 0.209 |
| 2 | 0.027 | 0.023 | 0.014 | 0.013 | 0.008 | 0.246 |
| 3 | 0.079 | 0.069 | 0.038 | 0.018 | 0.006 | 0.522 |

At the same time, the trajectory obtained from the relocation calculation is compared with the ground truth as shown in Figure 6.

We have applied special magnification processing to the point of maximum deviation between the estimated trajectory and the ground truth trajectory. It can be observed that, in terms of displacement error, the Z-axis shows a significant anomaly at the end of the sequence in the dense highway scene, reaching 0.7 meters. This is the reason why the average localization accuracy in this scene is noticeably lower than in the other two. At this moment, there are too many moving vehicles in the highway scene. Although our 3DGS mapping method can largely avoid motion blur during rendering, in extremely extreme situations, some blurring still occurs, affecting the accuracy of relocation.

To validate the sophistication of our method, we compare it with five representative relocation methods, which include DSAC++, AS, NG-RANSAC, Regression-only, and 3DGS-Reloc.DSAC++ is a deep learning-based relocation method, which estimates the camera position with high accuracy and generalization ability through differentiable RANSAC algorithm and CNN network to predict the scene coordinates and use PnP algorithm to estimate the camera position with high localization accuracy and generalization ability.AS (Active Search) is a feature-point matching based method that combines the F2P and P2F search mechanisms to optimize the matching process by using covariance relationship and solves the position by DLT and RANSAC algorithms, which is suitable for texture-rich scenes. NG-RANSAC is an improvement of the classical RANSAC algorithm, which optimizes the sampling strategy and computation process, and improves the operation efficiency and robustness. The Regression-only method predicts the camera position directly from the image by neural network, which does not rely on feature point extraction, and has fast computation and strong noise immunity. 3DGS-Reloc utilizes 3D Gaussian rendering technique for re localization, which is similar to the idea of our constructed method, but its requires additional sensors such as GPS provision to provide a rough initial position. These comparisons cover a wide range of techniques, from traditional feature matching to deep learning, and demonstrate the advantages of our approach.

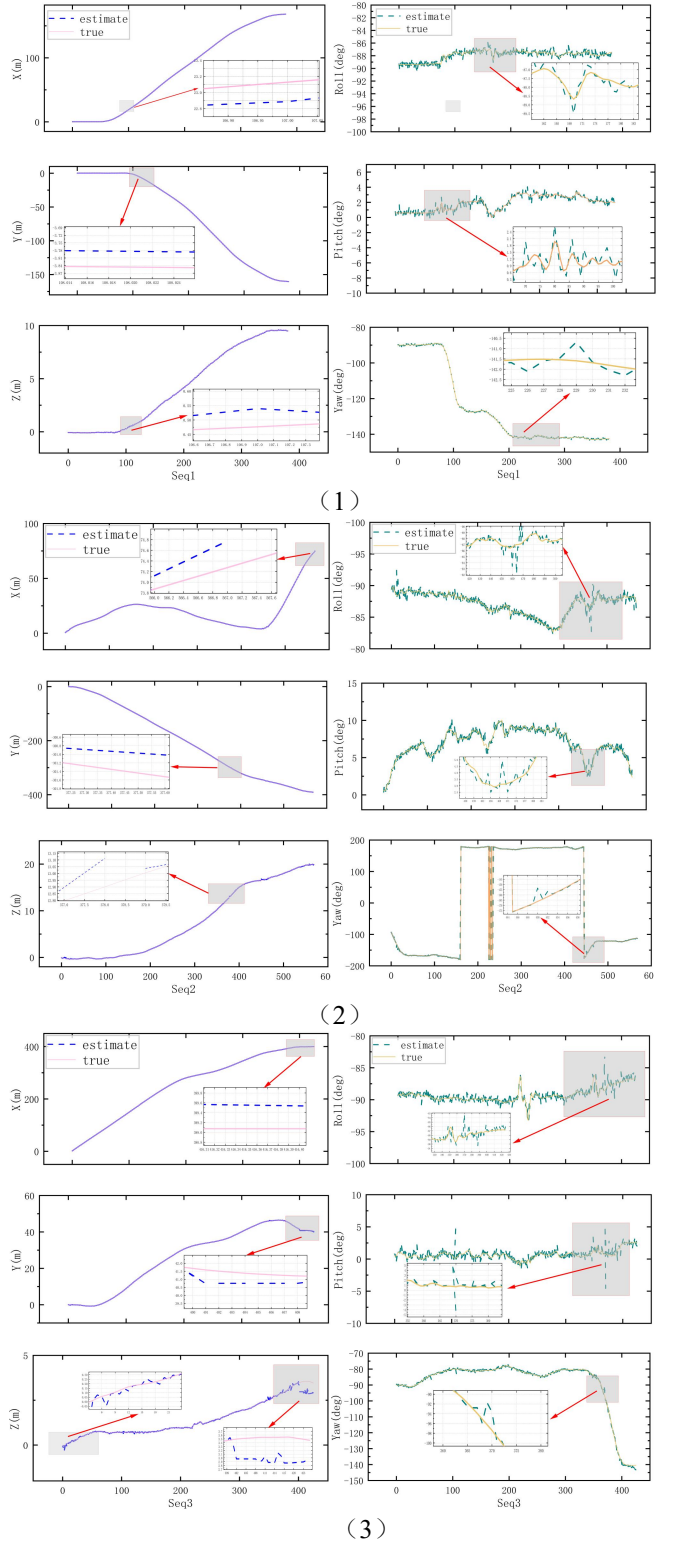

**Figure 6** Seq1, 2, and 3 predicted trajectories vs.ground truth

**Table 3:** Localization results on the KITTI, Waymo, and NuScenes datasets. The average absolute translation error (m) and absolute rotation error (°) are reported for each scene, and Acc. indicates the average accuracy for all scenes.

| Method | KITTI | | | Waymo | | | NuScenes | | |
|---|---|---|---|---|---|---|---|---|---|
| | Townroad | Boulevard | Avg | Townroad | Boulevard | Avg | Townroad | Boulevard | Avg |
| AS [41] | 16.6/0.75 | 21.2/0.84 | 18.9/0.79 | 15.1/0.77 | 19.2/0.71 | 17.1/0.74 | 16.3/0.59 | 20.5/0.62 | 18.4/0.61 |
| NG-RANSAC [42] | 13.9/- | 16.4/- | 15.2/- | 10.5/- | 11.6/- | 11.1/- | 8.1/- | 10.4/- | 9.25/- |
| Regression-only | 10.9/0.74 | 13.4/0.79 | 12.2/0.77 | 12.8/0.62 | 15.4/0.72 | 14.1/0.67 | 9.7/0.53 | 12.2/0.56 | 10.9/0.55 |
| DSAC++ [43] | 8.8/**0.67** | 10.2/0.72 | 9.5/0.70 | 7.8/0.53 | 11.3/0.63 | 9.6/0.58 | 6.5/**0.42** | 9.9/0.48 | 8.2/0.45 |
| 3DGS-Reloc [44] | 4.8/0.77 | 5.2/0.75 | 5.0/0.76 | **3.0**/0.64 | 4.2/0.71 | 3.6/0.68 | 4.4/0.53 | 3.8/0.57 | 4.1/0.55 |
| Ours | **2.9/0.67** | **3.5/0.71** | **3.2/0.69** | 3.1/**0.55** | **3.5/0.61** | **3.3/0.58** | **2.6/0.42** | **3.1/0.47** | **2.9/0.45** |

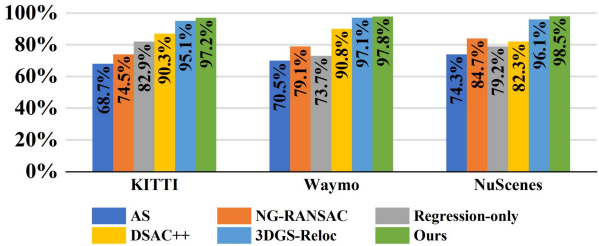

**Figure 7:** Percentage of High-Precision Relocalization Images

The results are reported in Table 3. Some results of competing methods were obtained from the respective papers, and others were derived from reproduced code. Overall, our method achieves excellent results. On the KITTI dataset, our method achieved localization errors of 2.9 cm/0.67° and 3.5 cm/0.71°, significantly outperforming the other five methods. On the Waymo and NuScenes datasets, our method also achieved the best overall localization performance. Although our method did not achieve the best results on all datasets in the Townroad scenario, it maintained excellent localization accuracy in the more challenging Boulevard scenario. We also reported the percentage of test images with localization errors less than 10 cm and 1 degree, as shown in Figure 7. This metric is used to comprehensively compare the localization performance of different methods. It can be seen that our method achieved the best performance across the three different datasets. Overall, our method achieved the lowest localization error in all scenarios and demonstrated good stability and adaptability across different environments. This indicates that our method has significant advantages in various complex environments.

### 4.4 Timeliness analysis

We analyzed the time consumption at each stage of the proposed model during the relocalization task. The relocalization process mainly includes the following steps: 1) querying the nearest anchor image from the retrieval database; 2) calculating the relative displacement between the query image and the nearest anchor image using the PnP algorithm; 3) rendering a new anchor image based on this relative displacement using a Gaussian model; 4) iteratively updating the localization result until the stopping condition is met.

**Table 4:** The time spent on each step during the relocalization process.

| stage | Running Speed | Number of Executions | | |
|---|---|---|---|---|
| | | Seq1 | Seq2 | Seq3 |
| Superpoint | 10ms/image | 25 | 35 | 45 |
| Superglue | 30ms/image | 25 | 35 | 45 |
| PnP | 2ms/image | 5 | 5 | 5 |
| render | 6ms/image | 5 | 5 | 5 |
| Time Spent | - | 1.04s | 1.44s | 1.84s |

In Table 4, we analyzed the time consumption at each stage and reported the relocalization time for three KITTI scenarios. The entire relocalization process can be completed within 2 seconds, indicating that this method can meet the real-time requirements of relocalization tasks in most autonomous driving scenarios and has the potential for practical application. In all scenarios, the iterative rendering converges in about five iterations. However, as the scene size increases, the number of anchor points increases, leading to a linear increase in retrieval time and a decrease in relocalization speed. By designing a reasonable anchor point selection strategy to reduce the size of the query database, the efficiency of the relocalization algorithm can be effectively improved.

## 5. Conclusion and discussion

In this paper, we construct a 3DGS map of outdoor large-scale environment based on the gassuion splatting technique of multi-sensor data, and combine the Superpoint and Superglue algorithms to accomplish efficient retrieval of sparse basemap library for RGB query images under large viewing angle deviation and reliable feature point matching, and utilize the rendering characteristics of 3DGS map to design the relocated iterative approximation structure to achieve High-precision relocation effect. We tested our method on numbers of sequences data from a variety of representative scenarios in the KITTI dataset, and all of them achieved centimeter-level localization accuracy, which is significantly improved compared to existing methods.

However, during the experiment, we found that because the multi-sensor data in the KITTI dataset is unidirectional trajectory data collected along the vehicle's driving route, the range of viewpoints from which the 3DGS maps constructed in this case can be rendered with high fidelity is limited, and thus if there is a significant deviation of the angle in the iterative process, it will result in a significant degradation of the quality of the 3DGS rendered maps, which will affect the convergence of the iterations. accuracy of the iterative convergence [45,46,47]. In future research, we can further explore how to make 3DGS maps rendered with high fidelity from a more free viewing angle in the case of limited data acquisition, and how to remove the sensitivity of the repositioning iteration process to angular errors.

## Funding Information

This research was supported by the National Natural Science Foundation of China under Grant No. 42374016.